%% file: main.tex
\documentclass[runningheads]{llncs}

\input{math_commands.tex}

\usepackage{graphicx}

\usepackage{tikz}
\usepackage{comment}
\usepackage{amsmath,amssymb} 
\usepackage{color}

\usepackage{epsfig}
\usepackage{url}
\usepackage[utf8]{inputenc} 
\usepackage[T1]{fontenc}    
\usepackage{subcaption}
\usepackage[normalem]{ulem} 
\usepackage[export]{adjustbox}
\usepackage{booktabs}       
\usepackage{amsfonts}       
\usepackage{nicefrac}       
\usepackage{microtype}      
\usepackage{gensymb}
\usepackage{tabularx}
\usepackage{siunitx}
\usepackage{enumitem}
\usepackage{lineno}
\usepackage{pifont}

\usepackage[font=small,labelfont=bf]{caption}

\newcommand{\ie}{\textit{i}.\textit{e}., }

\usepackage[accsupp]{axessibility}  

\begin{document}
\sloppy

\pagestyle{headings}
\mainmatter
\def\ECCVSubNumber{xxxx}  

\title{Active Audio-Visual Separation of\\ Dynamic Sound Sources} 

\titlerunning{Active Audio-Visual Dynamic Separation}
%
\author{Sagnik Majumder\inst{1}\index{Majumder, Sagnik}
\and
Kristen Grauman\inst{1, 2}\index{Grauman, Kristen}}
\authorrunning{Majumder et al.}
%
\institute{UT Austin, Austin, TX, USA 
\and Facebook AI Research, Austin, TX, USA\\
\email{\{sagnik, grauman\}@cs.utexas.edu}
}
\maketitle

\input{sections/abstract}

\input{sections/introduction}

\input{sections/related_work}

\input{sections/task}

\input{sections/approach}

\input{sections/experiments}

\input{sections/conclusion}

\clearpage
%
%
\bibliographystyle{splncs04}
\bibliography{mybib}

\clearpage
\input{sections/supp}
\end{document}

%% file: math_commands.tex
\usepackage{amsmath,amsfonts,bm}

\def\eqref#1{equation~\ref{#1}}

\def\1{\bm{1}}

\newcommand{\test}{\mathcal{D_{\mathrm{test}}}}

\DeclareMathAlphabet{\mathsfit}{\encodingdefault}{\sfdefault}{m}{sl}
\SetMathAlphabet{\mathsfit}{bold}{\encodingdefault}{\sfdefault}{bx}{n}



%% file: sections/abstract.tex
\begin{abstract}
We explore active audio-visual separation for dynamic sound sources, where an embodied agent moves intelligently in a 3D environment to \emph{continuously} isolate the \emph{time-varying} audio stream being emitted by an object of interest. The agent hears a mixed stream of multiple audio sources (e.g., multiple people conversing and a band playing music at a noisy party). Given a limited time budget, it needs to extract the target sound accurately at \emph{every step} using egocentric audio-visual observations. We propose a reinforcement learning agent equipped with a novel transformer memory that learns motion policies to control its camera and microphone to recover the dynamic target audio, using self-attention to make high-quality estimates for current timesteps and also simultaneously improve its past estimates. Using highly realistic acoustic SoundSpaces~\cite{chen2020soundspaces} simulations in real-world scanned Matterport3D~\cite{Matterport3D} environments, we show that our model is able to learn efficient behavior to carry out continuous separation of a dynamic audio target. Project: \url{https://vision.cs.utexas.edu/projects/active-av-dynamic-separation/}. 
\end{abstract}

%% file: sections/introduction.tex
\section{Introduction}\label{sec:intro}
\begin{figure}[t] 
    \centering
    \includegraphics[width=1.0\linewidth]{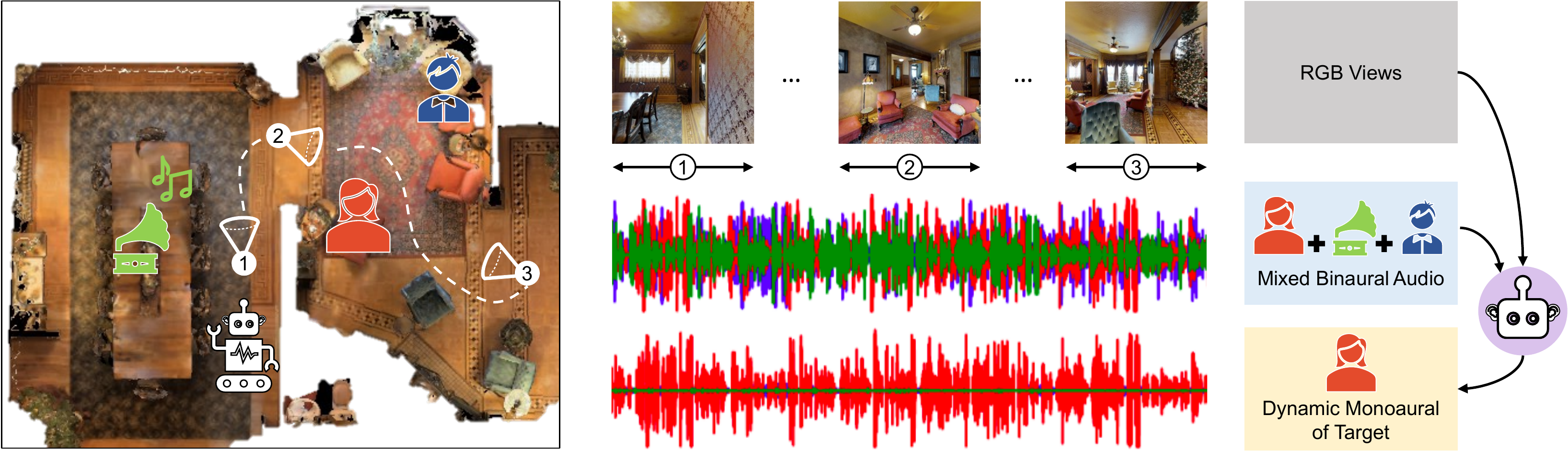} 
    \caption{ 
    Active audio-visual separation of dynamic sources. Given multiple \emph{dynamic} (time-varying, non-periodic) audio sources $S$, all mixed together, the proposed agent separates the audio signal by actively moving in the 3D environment on the basis of its egocentric audio-visual input, so that it is able to accurately retrieve the 
    target signal at every step of its motion.}
\label{fig:intro}
\end{figure}

Our daily lives are full of \emph{dynamic audio-visual events}, and the activity and physical space around us affect how well we can perceive them. For example, an assistant trying to respond to his boss's call in a noisy office might miss the initial part of her comment but then visually spot other noisy actors nearby and move to a quieter corner to hear better; to listen to a street musician playing a violin at a busy intersection, a pedestrian might need to veer around other onlookers and shift a few places to clearly hear.

These examples show how \emph{smart sensor motion} is necessary for accurate audio-visual understanding of dynamic events. For audio sensing in an environment full of distractor sounds, variations in acoustic attributes like volume, pitch, and semantic content of a sound source---in concert with a listener's proximity and direction from it---determine how well the listener is able to hear a target sound. However, just getting close or far is not always enough: effective hearing might require the listener to move around by \emph{visually sensing} competing sound sources and surrounding obstacles, and discovering locations favorable to listening (e.g., an intersection of two walls that reinforces listening through audio reflections, the front of a tall cabinet that can dull acoustic disturbances).

In this work, we investigate how to induce such intelligent behaviors in autonomous agents through  audio-visual learning. Specifically, we propose the new task of \emph{active audio-visual separation of dynamic sources}: given a stream of egocentric audio-visual observations, an agent must determine how to move in an environment with multiple sounding objects in order to \emph{continuously} retrieve the \emph{dynamic} (temporally changing) sounds being emitted by some object of interest, under the constraint of a limited time budget. See Figure~\ref{fig:intro}. This task 
is relevant for augmented reality (AR) and mobile robotics applications, where a user equipped with an assistive hearing device or a service robot needs to better understand a sound source of interest in a busy environment.

Our task is distinct from related efforts in the literature. Whereas traditional audio-visual separation models  extract sounds passively from pre-recorded videos~\cite{gabbay2017visual,gao2018learning,afouras2018conversation,ephrat2018looking,owens2018audio,gao2019co,chung2020facefilter,gao2021visualvoice,zadeh2019wildmix,chen2020dual,subakan2021attention,zhang2021transmask,lu2021spectnt}, our task requires active placement of the agent's camera and microphones over time. Whereas embodied audio-visual navigation~\cite{chen2020soundspaces,gan2019look,chen2020semantic,chen2021learning,anonymous2022sound} entails moving towards a sound source, our task requires recovering the sounds of a target object. The recent Move2Hear model performs active source separation~\cite{majumder2021move2hear} but is limited to \emph{static} (\ie periodic or constant) sound sources, such as a ringing phone or fire alarm---where recovering one timestep of the sound is sufficient. In contrast, our task involves \emph{dynamic} audio sources and calls for extracting the target audio at \emph{every step} of the agent's movement. Variations in the observed audio arise not only from the room acoustics and audio spatialization, but also from the temporally-changing and fleeting nature of the target and distractor sounds. This means the agent must recover a new audio segment at every step of its motion, which it hears only once.  The proposed task is thus both more realistic and more difficult than existing active audio-visual separation settings.

To address active dynamic audio-visual source separation, we introduce a reinforcement learning (RL) framework that trains an agent how to move to continuously listen to the dynamic target sound. Our agent receives a stream of egocentric audio-visual observations in the form of RGB images and mixed binaural audio, along with the target category of interest (human voice, musical instrument, etc.) and decides its next action (translation or rotation of its camera and microphones) at every time step. 

The technical design of our approach aims to meet the challenges outlined above. First, the proposed  motion policy accumulates its observations with a recurrent network.  Second, a transformer~\cite{vaswani2017attention}-based acoustic memory leverages self-attention to simultaneously produce an estimate of the current segment of the dynamic target audio while refining estimates of past timesteps. The bi-directionality of the transformer's attention mechanism helps  capture regularities in acoustic attributes (e.g., volume, pitch, timbre) and semantic content (words in a speech, musical notes, etc.) and model their variations over a long temporal range. Third, the motion policy is trained using a reward that encourages movements conducive to the best-possible separation of the target audio at every step, thus forcing both the policy and the acoustic memory to learn about patterns in the acoustic and semantic characteristics of the target.

We validate our ideas with realistic audio-visual 
simulations from SoundSpaces~\cite{chen2020soundspaces} together with 53 real-world Matterport3D~\cite{Matterport3D} environment scans, along with non-periodic sounds from many diverse human speakers, music, and other common background sources. Our agent successfully learns a motion policy for hearing its dynamic target more clearly in unseen scenes, outperforming  the state-of-the-art Move2Hear~\cite{majumder2021move2hear} approach 
in addition to several baselines.

%% file: sections/related_work.tex
\section{Related Work}\label{sec:related}

\paragraph{Audio(-Visual) Source Separation.}
Substantial prior work uses audio alone to separate sounds in a passive (non-embodied) way, whether from single-channel (monaural) audio~\cite{10.1007/978-3-540-74494-8_52,Spiertz09source-filterbased,Virtanen07monauralsound,6853860} or multi-channel audio that better surfaces spatial cues~\cite{nakadai2002real,Yilmaz04blindseparation,duong2010under,deleforge:hal-00768668,weiss2009source,zhang2017deep,gao2019visual-sound}.
Bringing vision together with audio, approaches based on signal processing and matrix factorization~\cite{hershey2000audio,NIPS2000_11f524c3,Smaragdis03audio/visualindependent,pu2017audio,7951787,7760220,gao2018learning}, visual tracking~\cite{4270342,7952688}, and deep learning~\cite{gan2020music,xu2019recursive,zhao2018sound,gao2019co} have all been explored, with particular recent emphasis on audio-visual separation of speech~\cite{afouras2018conversation,gabbay2017visual,ephrat2018looking,owens2018audio,afouras2019my,chung2020facefilter,gao2021visualvoice}. Other methods explore isolating a target audio source of interest~\cite{DBLP:conf/interspeech/OchiaiDKIKA20,tzinis2020into,DBLP:conf/interspeech/GuCZZXYSZ019,gu2020temporal,8736286}.

Active models for hearing better include sound source localization methods in robotics that steer a microphone towards a localized source  (e.g.,~\cite{asano2001,nakadai2000active,bustamante2018}) and audio-visual methods that detect when people are speaking~\cite{alameda2015vision,viciana2014audio,ban2018icassp,Ego4D2022CVPR}, which could help attend to one at a time.  Most recently, Move2Hear~\cite{majumder2021move2hear} trains an embodied agent to move in a 3D environment using audio and vision in order to hear a target object. Although it is an important first step in tackling active separation, Move2Hear deals only with static sounds, as discussed above. This severely limits its real-world scope, where almost all sounds, including those of interest like human speech, music, etc., are time-varying. Addressing dynamic sounds is decidedly more complex: the model must leverage not only the agent's surrounding geometry (to shut out undesirable interfering sounds) but also predict a continuously evolving audio sequence while hearing it just once within the mixed input sound. Aside from substantially generalizing the task, compared to~\cite{majumder2021move2hear} our core technical contributions are a novel transformer~\cite{vaswani2017attention}-based acoustic memory that does bi-directional self-attention to not only produce high-quality current separations on the basis of past separations but also refine past separations by taking useful cues from current separations, and a dense reward that promotes accurate dynamic separation at every step of the agent's motion. We demonstrate our model's advantages over~\cite{majumder2021move2hear} in results.

\paragraph{Transformer Memory.}
Transformers~\cite{vaswani2017attention} have been shown to do very well on long-horizon embodied tasks like navigation and exploration~\cite{fang2019scene,mezghani2020learning,chen2020semantic,chen2021topological,campari2020exploiting,mezghani2021memory,ramakrishnan2021environment}. The performance gains arise from a transformer's ability to effectively leverage past experiences of the agent~\cite{fang2019scene,mezghani2020learning,mezghani2021memory,campari2020exploiting,ramakrishnan2021environment} and do cross-modal reasoning~\cite{chen2020semantic,chen2021topological}. Different from these methods, our idea is to use a transformer as a memory model for capturing long-range acoustic correlations for audio-visual separation. Unlike prior approaches that show the successful use of transformers for passive separation~\cite{zadeh2019wildmix,chen2020dual,subakan2021attention,zhang2021transmask,lu2021spectnt}, we train an agent to actively position itself in a 3D environment so that the transformer memory can improve current and past estimates of the target audio through bi-directional self-attention.

%% file: sections/task.tex
\section{Task Formulation}

We introduce a novel embodied task: active audio-visual separation of dynamic sound sources. In this task, an autonomous agent hears multiple \emph{dynamic} audio sources of different types placed at various locations in an unmapped 3D environment, where the raw audio emitted by each source varies with time. The agent's objective is to move around intelligently to isolate the \emph{target source's}\footnote{e.g., a human speaker or instrument the agent wishes to listen to} dynamic audio from the observed audio mixture at every step of its motion. 

Our agent relies on both acoustic and visual cues to actively listen to the ever-changing target audio. 
The audio stream conveys cues about the spatial placement of the audio sources relative to the agent. Besides letting the agent see sounding objects within its field of view, the visual signal can help it avoid obstacles and go towards more closed spots in the 3D scene to suppress undesired acoustic interference. Both audio and vision can reveal how far the agent can venture in search of possibly favorable locations to cut out distractor sounds---without failing to follow the target. The temporal variations in the heard audio can also inform the agent about patterns in the fluctuations of the sound quality, intensity, and semantics (e.g., a voice raising and falling), letting it anticipate those attributes for future timesteps and move accordingly to maximize separation. 

\paragraph{Task Definition.} 
An episode of the proposed task begins by randomly instantiating multiple audio sources in a 3D environment by sampling their locations, types, and the dynamic audio tracks that play at the source positions. At every timestep, the agent receives a mixed binaural audio waveform, which depends on the surrounding scene's geometric and material composition, the relative spatial arrangement of the agent and the sources, and the audio types (e.g., speech, musical instrument, etc.). One of the sources is the target. The agent's goal is to extract the latent monaural signal playing at the target at every timestep.\footnote{The \emph{monaural} source is the ground truth target, since it is devoid of all material and spatial effects due to the environment and the mutual positioning of the agent and the sources.  Using a spatialized (e.g., binaural) audio as ground truth would permit undesirable shortcut solutions: the agent could move somewhere in the environment where the target is inaudible, and technically return the right answer of silence~\cite{majumder2021move2hear}.} 

To succeed, the agent needs to leverage egocentric audio-visual cues to move intelligently so that it can predict the dynamic audio sequence. The agent cannot re-experience sounds from past timesteps, but it is free to revise its past predictions.

\paragraph{Episode Specification.} 
We formally specify an episode by the tuple \break
$(\mathcal{E}, p_0, S_1(t), S_2(t), \dots S_k(t), G^c)$. Here, $\mathcal{E}$ refers to the 3D scene in which the episode is executed, $p_0=(l_0, o_0)$ denotes the initial agent pose determined by its location $l$ and orientation $o$, and $S_i(t)=(S_i^w(t), S_i^l, S_i^c)$ specifies an audio source by its dynamic monaural waveform $S_i^w$ as a function of the episode step $t$, location $S_i^l$ and audio type $S_i^c$. $k$ is the total number of audio sources of distinct types ($S_1^c\neq S_2^c \neq \dots \neq S_k^c$) present in the scene, and $G^c$ is the target audio type, such that $G\in\{S_i\}$. At each step, the agent's objective is to listen to the binaural mixture of all sources and predict the dynamic signal $G^w(t)$ associated with the target audio type $G^c$. Note that distinct human voices are considered distinct audio types. We limit the episode length to $\mathcal{T}$ steps, thus assigning the agent a fixed time budget to retrieve the entire audio clip. 

\paragraph{Action Space.}
At each timestep, the agent samples an action $a_t$ from its action space $\mathcal{A}=\{MoveForward, TurnLeft, TurnRight\}$ and moves on an unobserved \emph{navigability graph} of the environment. While turns are always valid, $MoveForward$ is not valid unless there is an edge from the current node to the next in the direction the agent is facing (i.e., no walls or obstacles exist there).

\paragraph{3D Environment and Audio-Visual Simulation.}
Following the state-of-the-art~\cite{chen2020soundspaces,chen2020semantic,majumder2021move2hear}  in embodied audio-visual learning, we use the AI-Habitat simulator~\cite{habitat19iccv}  with SoundSpaces~\cite{chen2020soundspaces}  audio simulations and Matterport3D scenes~\cite{Matterport3D}. Matterport3D provides dense 3D meshes and image scans of real-world houses and other indoor environments. SoundSpaces contains room impulse responses (RIR) to render realistic spatial audio at a resolution of 1 meter for Matterport3D. It  models how sound waves from each source travel in the 3D environment and interact with the surrounding materials and geometry. In particular, the state-of-the-art RIRs simulate most real-world acoustic phenomena: direct sounds, early specular/diffuse reflections, reverberations, binaural spatialization, and frequency-dependent effects of materials and air absorption.  See~\cite{chen2020soundspaces} for details. By using these
highly perceptually realistic simulators together with real-world scanned environments and real-world audio data (cf.~Sec.~\ref{sec:experiments}), we minimize the sim2real gap; we leave exploring transfer to a real robot for future work.

In every episode, we place $k$ point objects as the monaural sources in a Matterport 3D scene and render a binaural mixture $B^{mix}_t$ of the audio coming to the agent from  all the source locations, by convolving the monaural waveforms at each step with the corresponding RIRs and taking their mean. Our agent moves through the 3D spaces while receiving real-time egocentric visual and audio observations that are a function of the 3D scene and its surface materials, the agent's current pose, and the sources' positions.

\begin{figure*}[t] 
    \centering
    \includegraphics[width=0.8\linewidth]{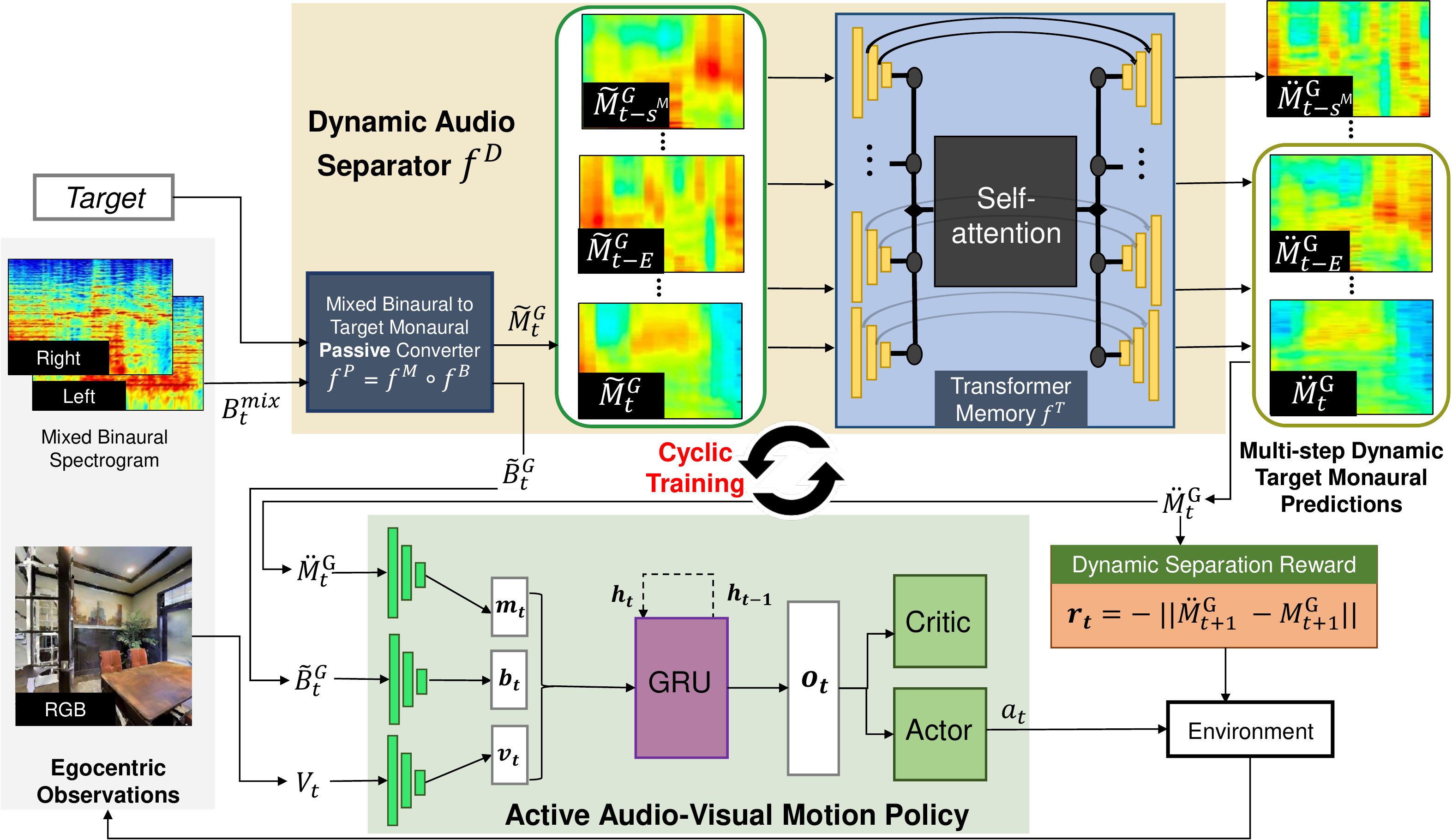} 
    \caption{
Our model addresses active audio-visual separation of dynamic sources by leveraging a synergy between a \emph{dynamic audio separator} and an \emph{active audio-visual motion policy}. The dynamic separator $f^D$ uses self-attention to continuously isolate a target signal $M^G_t$ from its received mixed binaural $B^{mix}_t$ on the basis of its past and current initial estimates $\{\tilde{M}^G_{t-s^\mathcal{M}}, \ldots, \tilde{M}^G_t\}$, while also using its current initial separation $\tilde{M}^G_t$ to refine past final separations $\{\ddot{M}^G_{t-E}, \ldots, \ddot{M}^G_{t-1}\}$. The motion policy uses the separator's outputs, $\ddot{M}^G_t$ and $\tilde{B}^G_t$ and egocentric RGB images $V_t$ to guide the agent to areas suitable for 
separating future dynamic targets.} \label{fig:model}
\end{figure*}

%% file: sections/approach.tex
\section{Approach}\label{approach}
We approach the problem with reinforcement learning. We train a motion policy to make sequential movement decisions on the basis of egocentric audio-visual observations, guided by dynamic audio separation quality. Our model has two main components (see Fig.~\ref{fig:model}): 1) an audio separator network and 2) an active audio-visual (AV) motion policy.

The separator network serves three functions at every step: 1) it passively separates the current target audio segment from its heard mixture, 2) it improves its past separations by exploiting their correlations in acoustic attributes like volume, pitch, quality, content semantics, etc. with the current separation, and 3) it uses the current separation to guide the motion policy towards locations favorable for accurate separation of future audio segments. The motion policy is trained to repeatedly maximize dynamic separation quality.  It moves the agent in the 3D scene to get the best possible estimate of the \emph{complete} target signal. 

These two components share a symbiotic relationship, each feeding off of useful learning cues from the other while training.  This allows the agent to learn the complex links between separation quality and the dynamic acoustic attributes of the target source, its location relative to the agent, the scene's material layout (walls, floors, furniture, etc.), and the inferred spatial arrangement of distractor sources in the 3D environment.

\subsection{Dynamic Audio Separator Network}
\label{approach:audio_net}
The dynamic audio separator network $f^D$ receives the mixed binaural sound $B^{mix}_{t}$ coming from all sources in the environment and the target audio type $G^c$ at every step. Using these inputs, it produces estimates $\ddot{M}^G$ of the monaural target at both the current and earlier $E$ steps,
i.e.,  $f^D(B^{mix}_t, G^c) = \{\ddot{M}^G_{t-E}, \ldots, \ddot{M}^G_t\}
$ (Fig.~\ref{fig:model} top). We use short-time Fourier transform (STFT) to represent the monaural $M$ and binaural $B$ as audio magnitude spectrograms of dimensionality $F \times N$ and $2 \times F \times N$, respectively, where $F$ is the number of frequency levels, $N$ is the number of overlapping temporal windows, and $B$ has 2 channels (left and right).   

The audio separation takes place by using two modules, such that $f^D = f^P \circ f^T$. First, the passive audio separator module $f^P$ predicts the current monaural target $\tilde{M}^G_t$, given the audio goal category $G^c$ and the binaural mixture $B^{mix}_t$. Next, the transformer memory $f^T$ uses self-attention to combine cues from the current monaural prediction $\tilde{M}^G_t$ and an external memory bank  $\mathcal{M}$ of $f^P$'s past predictions to both produce an enhanced version $\ddot{M}^G_t$ of the current monaural prediction and also refine the past $E$ predictions $\{\ddot{M}^G_{t-E}, \ldots, \ddot{M}^G_{t-1}\}$. Our multi-step prediction is motivated by the time-varying and fleeting nature of dynamic audio; we leverage semantic and acoustic structure in the target signal to both anticipate future targets and correct errors in past estimates. Next, we describe each of these modules in detail.

\paragraph{Passive Audio Separator.}
We adopt a factorized architecture~\cite{majumder2021move2hear} for the passive separator $f^P$, which comprises a binaural extractor $f^B$ and a monaural predictor $f^M$. Given the mixed binaural and the target audio type, $f^B$ predicts the target binaural $\tilde{B}^G_t$ by estimating a real-valued ratio mask~\cite{gao2019co,xu2019recursive,zhao2018sound}. $f^M$ converts $B^G_t$ into an initial estimate $\tilde{M}^G_t$ of the current monaural target. Both $f^B$ and $f^M$ use U-Net like architectures~\cite{RFB15a} with ReLU activations. We enhance these initial monaural predictions with a transformer memory, as we will describe next.

\paragraph{Transformer Memory.} 
The proposed memory is a transformer encoder~\cite{vaswani2017attention} that encodes the current and past $s^\mathcal{M}$ monaural predictions from $f^M$. Every transformer block has a pre-norm architecture that has been found to work particularly well for audio separation~\cite{subakan2021attention,zhang2021transmask}. For storing past estimates from $f^M$, we maintain an external memory bank of size $s^{\mathcal{M}}$. At every step, we encode the past and current estimates using a multi-layer convolutional network to output a lower-dimensional monaural feature set $\{e_{t-s^{\mathcal{M}}}, \ldots, e_{t}\}$. We feed these monaural features to the transformer along with sinusoidal positional encodings. 

The transformer encoder uses the monaural features for its keys, values, and queries to build a joint representation through self-attention, such that all features across the temporal range $[t-s^{\mathcal{M}}, t]$ can share information with each other. Further, the positional encoding for every feature informs the transformer about the feature's temporal position in its context, thus helping it capture both short- and long-range similarities in the  target dynamic signal.

On the transformer's output side, we sample the feature set $\{d_{t-E}, \ldots,d_t\}$ associated with the timesteps in the range $[t - E, t]$, where $E \leq s^{\mathcal{M}}$, and decode them using another multi-layer convolution to obtain monaural estimates $\{\ddot{M}^G_{t-E}, \ldots, \ddot{M}^G_t\}$.  We also connect the convolutional encoder and decoder with additive skip connections for faster training and improved generalization~\cite{he2016deep}.

This multi-step prediction is particularly useful in the dynamic setting, where at each step, the agent hears a new segment of the target sound. Consequently, the agent can benefit from an error-correction mechanism that uses the current estimates of the target to amend for mistakes in past estimates. Our results show that unlike RNN-based alternatives, the transformer memory naturally provides the agent with this option as it can exploit temporally local and global patterns in the dynamic audio---such as regularities in pitch and volume in human speech, and notes and timbre in music---in a bi-directional manner. By doing self-attention on an explicit storage of memory entries in a non-sequential fashion, it can accurately separate the current audio target and also refine past separations. The high quality of the current separations allows the agent to search for separation-friendly locations in its environment more extensively, by providing robustness in separation even when the agent temporarily passes through areas with high distractor interference.

Both modules $f^P$ and $f^T$ are trained in a supervised manner using the ground truth of the target binaural and monaural signals, as detailed in Sec.~\ref{approach:training}.

 \subsection{Active Audio-Visual Motion Policy}
\label{approach:policy}

The audio-visual (AV) motion policy guides the agent in the 3D environment to predict the best possible estimate of the dynamic audio target at every timestep (Fig.~\ref{fig:model} bottom). On the basis of real-time egocentric visual and audio inputs, the AV motion policy sequentially emits actions $a_t$ that help continuously maximize the separation quality of the dynamic separator network $f^D$ (Sec.~\ref{approach:audio_net}). It has two modules: an observation encoder and a policy network.

\paragraph{Observation Space and Encoding.}
At each step, the AV motion policy receives the egocentric RGB image $V_{t}$, the current binaural separation $\tilde{B}^G_t$ from $f^B$, and the current monaural prediction $\ddot{M}^G_t$ from $f^T$.

The audio and visual streams provide the agent with useful complementary information for continuous separation of the dynamic target. $\tilde{B}^G$ informs the agent about the relative spatial position (both distance and direction) of the target and also the scene's materials and geometry, allowing it to anticipate how these factors might affect its future movements and separation quality. The binaural cue also prevents the agent from venturing too far away from the target, where it would be unable to retrieve the current dynamic target, even with the help of its transformer memory. Besides informing the agent about the relation between separation quality and its pose relative to the target, as implicitly inferred from $\tilde{B}^G$, $\ddot{M}^G$ also allows the agent to track acoustic and semantic similarities in $\ddot{M}^G$ over time by using its recurrent memory module (see below) and choose an action that could help maximize the next separation.

The visual signal $V_t$ informs the policy about the 3D scene's geometric configuration and helps it avoid collisions with obstacles. The synergy of vision and audio lets the agent learn about relations between the 3D scene's geometry, semantics, and surface materials on the one hand, and the expected separation quality for different agent poses on the other hand.

We learn three separate CNN encoders to represent these inputs: $v_{t} = \mathcal{F}^{V}(V_{t})$, $b_{t} = \mathcal{F}^B(\tilde{B}^G_t)$ and $m_{t} = \mathcal{F}^{M}(\ddot{M}^G_t)$.
We concatenate $v_t, b_t \text{ and } m_t$ to obtain the complete audio-visual representation $o_t$.

\paragraph{Policy Network.}\label{para:policy}
The policy network in our model is a gated recurrent neural network (GRU)~\cite{chung2015recurrent} followed by an actor-critic architecture. The GRU receives the current audio-visual representation $o_t$ and its accumulated history of states from the last step $h_{t-1}$ to produce an updated history $h_t$ and also emit the current state feature $s_t$. The actor-critic module receives $s_t$ and $h_{t-1}$, and predicts the policy distribution $\pi_{\theta}(a_{t} | s_{t}, h_{t-1})$ and the current state's value  $\mathcal{V}_{\theta}(s_{t}, h_{t-1})$, where $\theta$ are the policy parameters. The agent samples actions $a_t$ from $\mathcal{A}$ as per the policy distribution to interact with the 3D environment.

\subsection{Training}
\label{approach:training}

\paragraph{Dynamic Audio Separator Training.}
As discussed in Sec~\ref{approach:audio_net}, the separator network has two modules: the passive separator $f^P$ that outputs $\tilde{B}$ and $\tilde{M}$, and the transformer memory $f^T$ that outputs $\ddot{M}$. We train both modules using the respective target spectrogram ground truths ($B^G$ for binaural and $M^G$ for monaural), which are provided by the simulator.

We train $f^P$ with the passive separation loss:
\begin{equation}\label{eq:loss_passive}
    \mathcal{L}^P_t =||\tilde{B}^G_t - B^G_t||_{1} + ||\tilde{M}^G_t - M^G_t||_{1}.
\end{equation}
The training loss $\mathcal{L}^T_t$ for the transformer memory aims to maximize the quality of both past and current monaural estimates, $\{\ddot{M}^G_{t-E},\ldots, \ddot{M}^G_{t-1}\}$ and $\ddot{M}^G_t$, respectively:
\begin{equation}\label{eq:loss_transformer}
    \mathcal{L}^T_t = \frac{1}{E + 1}\sum_{u=t-E}^{t}{||\ddot{M}^G_u - M^G_u||_{1}}.
\end{equation}
Note $f^P$ outputs step-wise predictions ($\tilde{B}^G_t$ and $\tilde{M}^G_t$), unlike $f^T$ that uses past separation history to produce its separation output for the current step and its current output to correct past errors. Hence, we pretrain $f^P$ using $\mathcal{L}^P$ and a static dataset pre-collected from the training scenes by randomly instantiating the number of audio sources, their locations, the target audio type, the static monaural segments for the sources, and the agent pose for each data point. Besides reducing the online computation cost by lowering the amount of trainable parameters, preliminary experiments showed that pretraining leads to stationarity in the distribution of $\tilde{M}^G$
during the early phases of on-policy updates, thus stabilizing the training of the transformer. 

Having pretrained $f^P$, we jointly train $f^T$ and the AV motion policy with on-policy data while keeping the parameters of $f^P$ frozen, since the agent's trajectories directly affect the distribution of inputs for both $f^T$ and the policy and hence, their training.

\paragraph{Audio-Visual Motion Policy Training.}
The dynamic separation task calls for a continuous retrieval of the time-varying target audio by the agent. Towards this goal, we train our motion policy with a dense RL reward: 
\begin{equation}
    r_t = -||\ddot{M}^G_{t+1} - M^G_{t+1}||_{1}.
\end{equation}
At each step, $r_t$ encourages the agent to act to maximize the next separation.

We train the policy using Decentralized Distributed PPO (DD-PPO)~\cite{wijmans2019dd} with trajectory rollouts of 20 steps. The DD-PPO loss consists of a value loss, policy loss, and entropy loss to promote exploration (see Supp. Sec.~\ref{sec:supp_train_hyperparams}).

\paragraph{Joint Training.}
We train the dynamic separator and AV motion policy by switching training between them after every parameter update.

%% file: sections/experiments.tex
\section{Experiments}
\label{sec:experiments}

\paragraph{Experimental Setup.}

We instantiate each episode with $k=2$ audio sources that are placed randomly in the scene at least 8 m apart and specify one source as the target (Supp.~details the dataset construction (Sec.~\ref{sec:supp_audio_data}) and shows results for $k > 2$ (Sec.~\ref{sec:supp_num_sources})). To clearly distinguish our task from audio-navigation, the agent is initially co-located with the target source at the episode start and is expected to fine-tune its position to isolate the dynamic audio target. We set the time budget $\mathcal{T}$ to 20 for all episodes. We split 53 large Matterport3D scenes into non-overlapping train/val/test with 731K/100/1K episodes; we always  evaluate the agent in unseen, unmapped environments.

Our dataset consists of 102 distinct types of sounds from three broad categories: speech, music, and background sounds. For speech, we use 100 different speakers from LibriSpeech~\cite{7178964}. For music, we combine multiple instruments from the MUSIC~\cite{zhao2018sound} dataset. For background sounds, we use non-speech and non-music sounds from ESC-50~\cite{piczak2015dataset}, like running water, dog barking, etc. One of the 100 speakers (all speakers are considered distinct types) or music is the target. The distractors can be either background sounds, a speaker, or music. All sources play unique audio types. Our diverse dataset lets us evaluate different separation scenarios of varying difficulty: speech vs.~speech, speech vs.~music, or subtracting assorted background sounds.

There are 3,292 clips in total across all categories for use as monaural audio, each at least 20 s long. This helps avoid repetition in the monaural audio at each source in an episode with $\mathcal{T}=20$. For unheard sounds, we split the clips into non-overlapping train/val/test in 18:3:4. For pretraining $f^P$, we generate 200K 1 s segments from the long clips and split it into train/val in 47:3, where the segments for each split are sampled from the corresponding split for long clips.

\paragraph{Existing method and baselines.}
We compare against the following methods:
\begin{itemize}[leftmargin=*,topsep=0pt,partopsep=0pt,itemsep=0pt,parsep=0pt]
\item\textbf{Stand In-Place:} audio-only agent that does not change its pose throughout the episode, thereby separating the target in a completely passive way.
\item\textbf{Rotate In-Place:} audio-only agent that rotates clockwise at its starting location at all steps to sample the mixed audio from all possible directions
\item\textbf{DoA:} audio-only agent that takes one step away from the target, turns around, and samples the direct sound 
by facing its direction of arrival (DoA)~\cite{nakadai2000active} until the episode ends.
\item\textbf{Random:} 
audio-only agent that randomly selects actions from the action space $\mathcal{A} $ at all steps.
\item\textbf{Proximity Prior:} an agent that selects random actions but stays inside a radius of 2 m (selected through validation) of the target so 
that it does not lose track of the dynamic goal audio. Unlike our agent, this agent assumes access to the privileged information of the ground truth distance to the target.
\item\textbf{Novelty~\cite{bellemare2016unifying}:} a visual exploration agent  trained with RL to maximize its coverage of novel locations in the environment within the episode budget.  This helps sample diverse audio cues which can potentially aid dynamic separation.
\item\textbf{Move2Hear~\cite{majumder2021move2hear}:} a state-of-the-art audio-visual RL agent for active source separation that was originally proposed to tackle static audio sources. We retrain this model on our task by using the dynamic separation reward (Sec.~\ref{approach:training}). 
\end{itemize}

For fair comparison, we use the same dynamic audio separator $f^D$ for all baselines and our agent, while the acoustic stream that the separator for each individual model receives depends on the its agent's actions. While $f^P$ shares its pre-trained parameters across all agents, $f^T$ is trained separately for each agent by collecting on-policy data as per the individual agent's policy. This helps disentangle the contributions to the separation quality stemming from the separator network and the policy designs, respectively. We set $s^\mathcal{M}$ to 19 and $E$ to 14 on the basis of validation (see Sec.~\ref{sec:supp_E_effect} for additional analysis on the choice of $E$). These choices allow $f^T$ to do accurate separation by leveraging the maximum amount of context possible in an episode ($s^\mathcal{M} = \mathcal{T}-1$) while also refining past predictions. 
This way there is a balance between the amount of forward and backward information flow through $f^T$'s bi-directional self-attention.

\paragraph{Evaluation.}
We evaluate all models in terms of mean separation quality of the dynamic monaural target over all $\mathcal{T}$ steps, averaged over 1,000 test episodes and 3 random seeds. We use standard metrics: \textbf{STFT distance}, the error between separated and ground truth spectrograms,
and \textbf{SI-SDR}~\cite{8683855}, a scale-invariant measure of the amount of distortion present in the separated waveform. 

We provide more details about the data, spectrogram computation, baseline implementation, network architectures, training and evaluation in Supp Sec.~\ref{sec:supp_audio_data}-\ref{sec:supp_sep_metrics}.

\subsection{Source Separation Results}
Table~\ref{sub_table:2source_main} shows the separation quality of all models. The passive baselines Stand and Rotate In-Place fare worse than the ones that move and sample more diverse audio cues, like Random and Proximity Prior. However, despite being able to move, Novelty~\cite{bellemare2016unifying} performs poorly; in its effort to maximize coverage of the environment, it wanders too far from the target and fails to hear it in certain phases of its motion. DoA improves over the stationary baselines, since standing a step away from the target source allows it to sample a cleaner audio cue.

On the unheard sounds setting, Move2Hear~\cite{majumder2021move2hear} outperforms some baselines (e.g., DoA, Random, Novelty~\cite{bellemare2016unifying}, etc.) but fares comparably against others (e.g., stationary baselines and Proximity Prior). 
This behavior can be attributed to the absence of the transformer memory $f^T$ in Move2Hear. Although it is trained to maximize the separation quality at every step, its acoustic memory refiner is unable to handle the time-varying and transient nature of audio. To further illustrate our task complexity, we plot the distribution of separation performance (SI-SDR) of Move2Hear for both static and dynamic sources in Fig.~\ref{fig:pr_vs_nnpr}. The switch from dynamic to static audio results in a substantial degradation of separation quality (compare dotted and solid red curves: the mode of the SI-SDR distribution shifts leftward upon going from dynamic to static). This clearly shows the distinct new challenges posed by the dynamic source separation setting.

\begin{table}[!t]
           \centering
            \centering
            \caption{Active audio-visual dynamic source separation.}
              \scalebox{0.75}{
              \setlength{\tabcolsep}{8pt}
                \begin{tabular}{l cc|cc}
                \toprule
                &   \multicolumn{2}{ c| }{\textit{Heard}} &  \multicolumn{2}{ c }{\textit{Unheard}}\\
                Model  & {SI-SDR $\uparrow$} & {STFT $\downarrow$} & {SI-SDR $\uparrow$} & {STFT $\downarrow$}\\
                \midrule
                Stand In-Place      & 2.49  & 0.328 & 2.03  & 0.343 \\
                Rotate In-Place     & 2.50  & 0.327 & 2.04  & 0.343  \\
                DoA               & 2.78  & 0.313 & 1.88  & 0.342 \\
                Random             & 2.81  & 0.314 & 1.95  & 0.343 \\
                Proximity Prior    & 2.92  & 0.309 & 2.05  & 0.338\\ 
                Novelty~\cite{bellemare2016unifying} & 1.68  & 0.358 & 1.44  & 0.366 \\ 
                Move2Hear~\cite{majumder2021move2hear} & 2.31 & 0.331 & 2.06 & 0.339 \\
                \midrule
                Ours & \textbf{3.93} & \textbf{0.273} & \textbf{2.57} & \textbf{0.318}\\
    \midrule
    Ours w/o transformer memory $f^T$      & 2.32  & 0.330 & 2.02  & 0.340 \\
    Ours w/o multi-step predictions (\ie $E=0$)                & 2.69 & 0.317 & 2.29 & 0.33 \\
    Ours w/o visual input $V_t$ & 3.61 & 0.284 & 2.40 & 0.324 \\
                \bottomrule
              \end{tabular}
              }
              \label{sub_table:2source_main}
\end{table}

Our model outperforms all baselines and Move2Hear by a statistically significant margin  ($p \leq 0.05$). It overcomes the drop in Move2Hear's separation quality induced by dynamic sources (compare solid red and green curves in Fig.~\ref{fig:pr_vs_nnpr}: the 
mode for dynamic audio shifts rightward upon using our method). Its performance demonstrates the impact of our active policy and long-term memory. Simply staying at or close to the source to be able to continuously hear the target (as done by Stand or Rotate In-Place, DoA, and Proximity Prior) is not enough. Our improvement over Move2Hear~\cite{majumder2021move2hear} further emphasizes the advantage of our transformer memory $f^T$ in dealing with dynamic audio, both in terms of boosting separation when the agent is able to sample a cleaner signal, and providing robustness to the separator when the agent is passing through zones that are relatively less suitable for separation.

\subsection{Model Analysis}\label{subsec:model_analysis}
\paragraph{Ablations.} 
We report the ablation of our model components in Table~\ref{sub_table:2source_main} bottom. Our model suffers a severe drop in performance upon removing our transformer memory ($f^T$). This shows that $f^T$ plays a significant role in boosting dynamic separation quality. $f^T$ learns joint representations of past and present separation estimates through self-attention, successfully leveraging long-range temporal relations in the audio signal for high-quality separation. The drop when $E$ is set to 0 shows that the proposed multi-step predictions are also important in the dynamic setting, where the agent hears each segment of the time-varying signal only once and might not be able to separate a target very well at its first attempt. Our multi-step predictions let the agent improve past separations over time by extrapolating acoustic and semantic cues present in the current estimate. The visual cue $V_t$ also contributes to our agent's separation performance;  without it, the agent is prone to collisions and has to solely rely on the separated binaural $B^G_t$ to understand how the surrounding geometry and materials affect separation.

\paragraph{Transformer Memory.}\label{para:transformer_analysis}

\begin{figure}[t]
    \centering
    \begin{subfigure}[b]{0.32\linewidth}
    \centering
    \includegraphics[width=\linewidth]{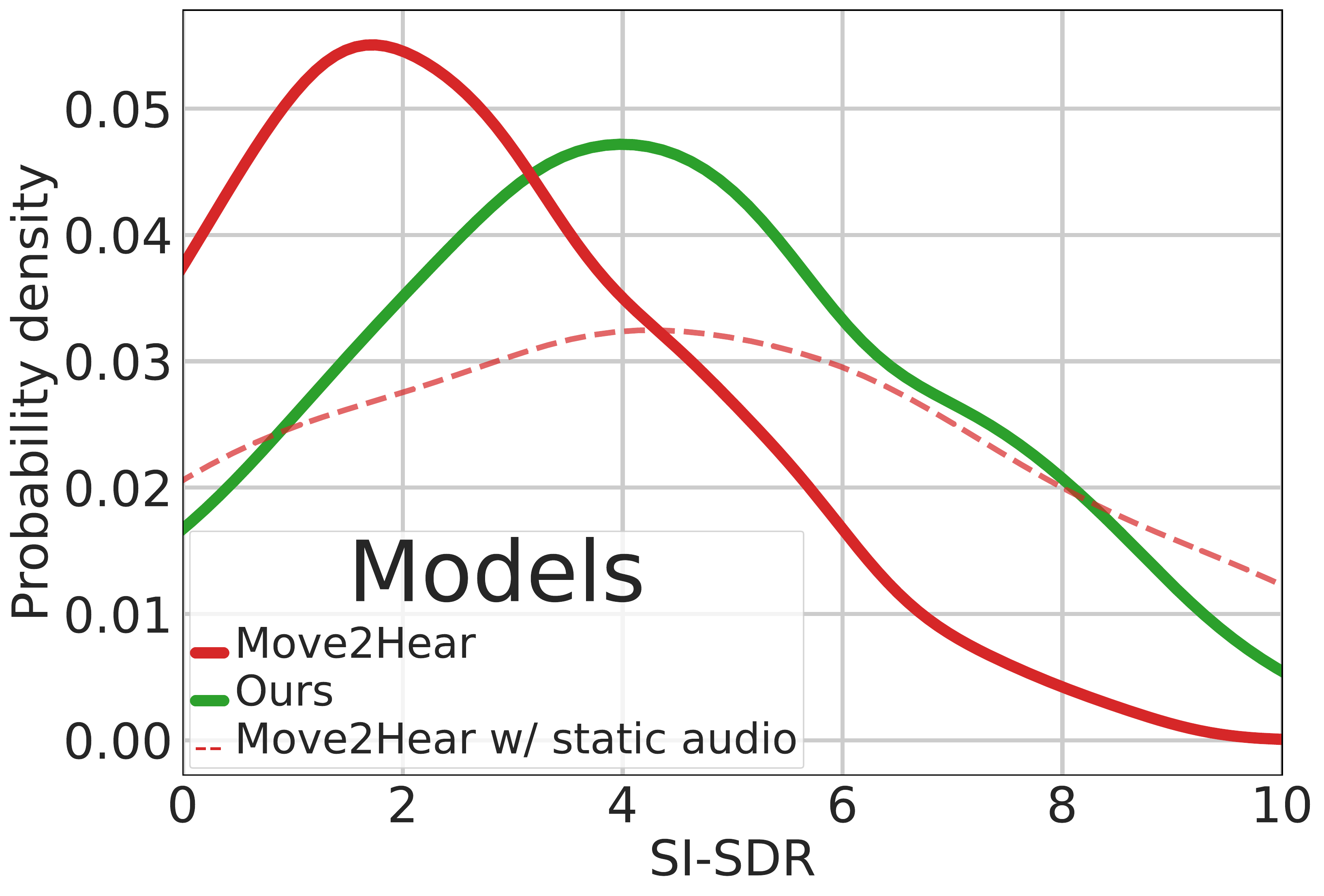}
    \caption{Separation quality distribution for static and dynamic audio}
    \label{fig:pr_vs_nnpr}
    \end{subfigure}\hfill
    \begin{subfigure}[b]{0.32\linewidth}
    \centering
    \includegraphics[width=\linewidth]{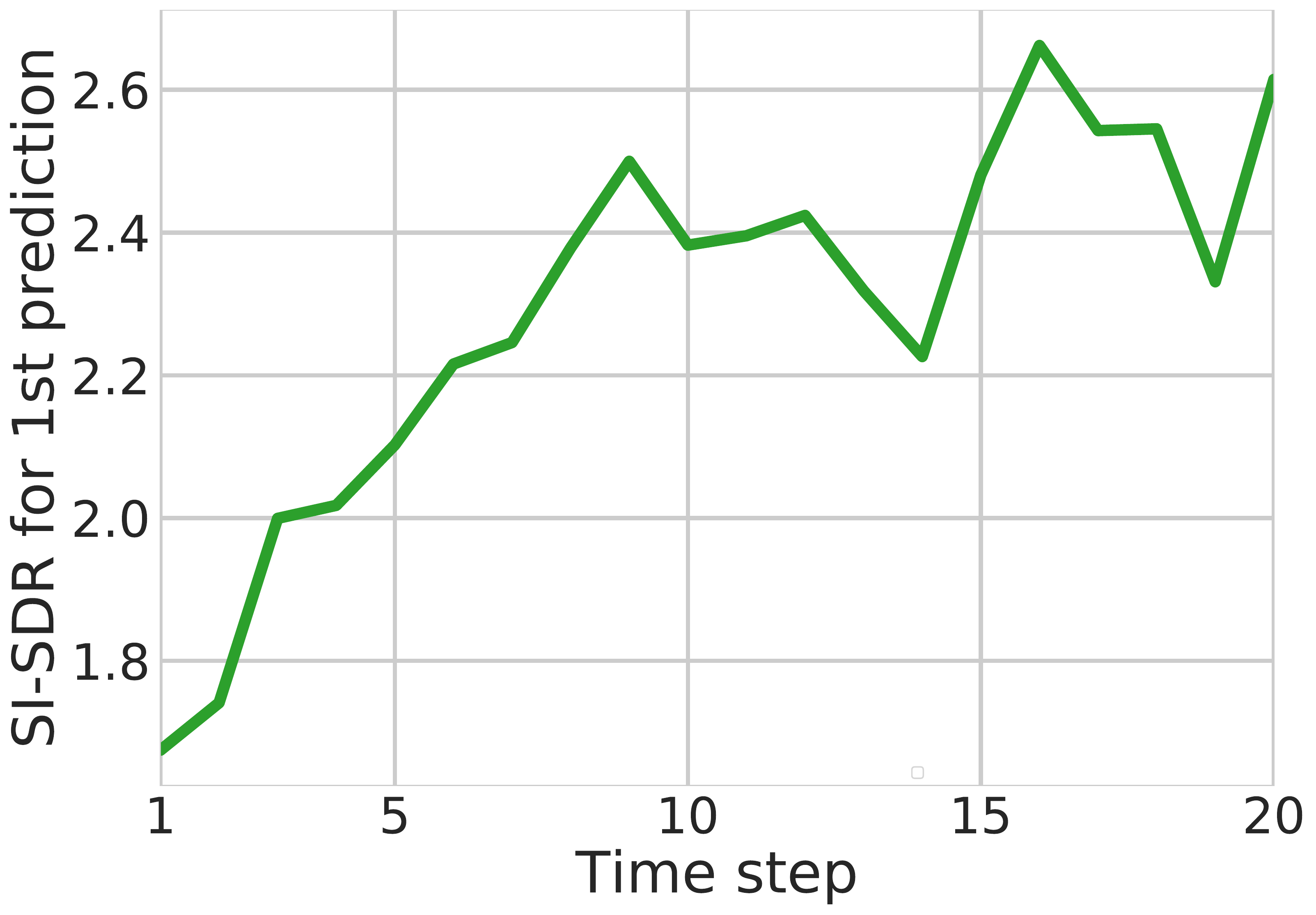}
    \caption{First separation with episode step}
    \label{fig:first_sep_vs_time}
    \end{subfigure}\hfill
    \begin{subfigure}[b]{0.32\linewidth}
    \centering
    \includegraphics[width=\linewidth]{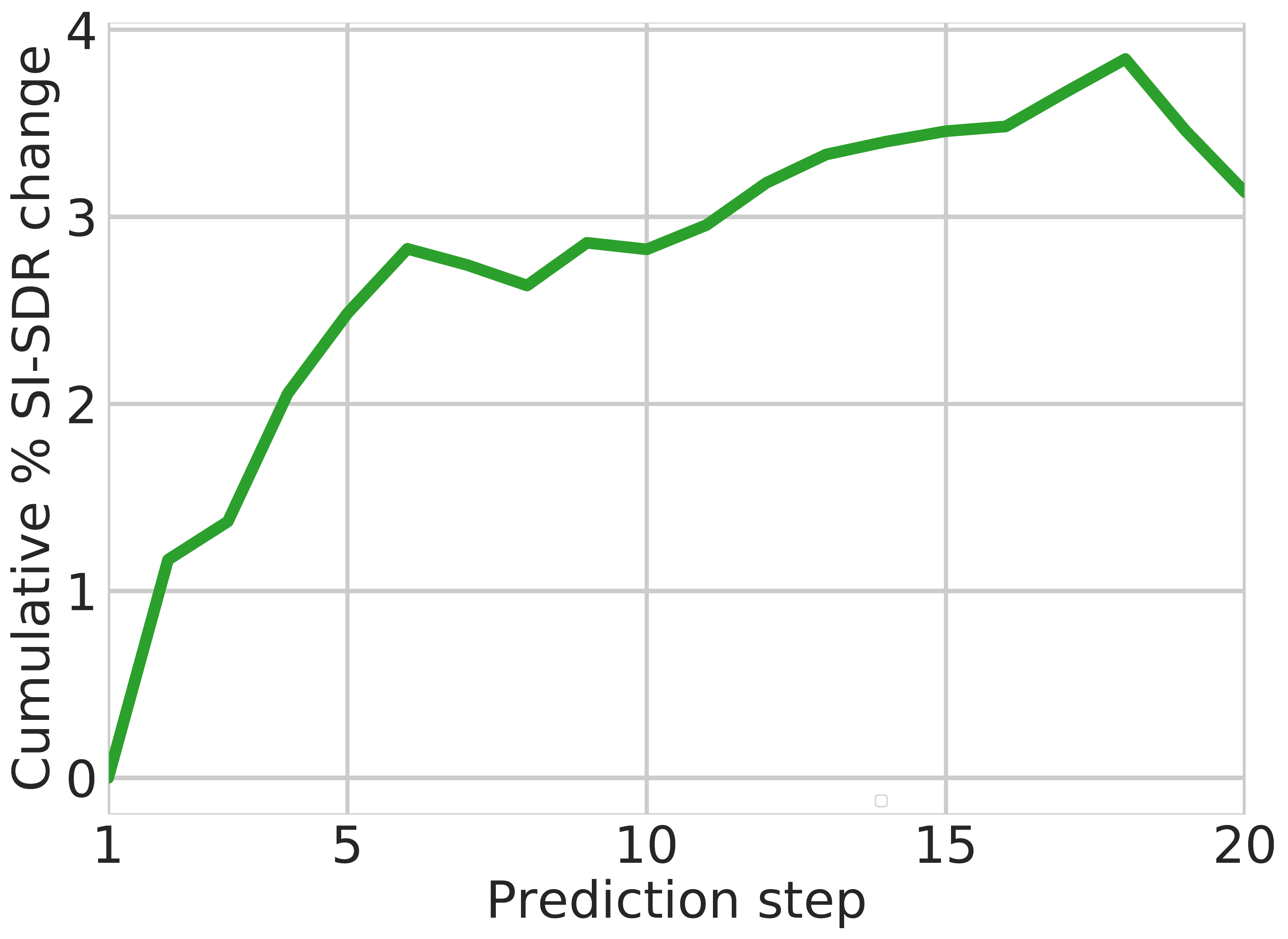}
    \caption{Separation with  prediction count}
    \label{fig:sep_vs_num_preds}
    \end{subfigure}
\caption{(a) Distribution of separation quality for static and dynamic audio (higher SI-SDR is better). (b) Quality at first separation as a function of time. (c) Separation quality for a target as a function of prediction count.}
    \label{fig:steps_k3}
\end{figure}

Next we analyze different aspects of our transformer memory $f^T$ to understand how they boost our agent's separation performance. 

Fig.~\ref{fig:first_sep_vs_time} shows how a longer memory leads to better initial estimates of the target audio segments. With more memory, $f^T$ is better able to find consistencies in past estimates to improve the current estimate.

Fig.~\ref{fig:sep_vs_num_preds} shows how making more estimates of the same target using the multi-step prediction mechanism of $f^T$ improves separation. Our model keeps refining all its predictions until late in the episode. This shows how $f^T$ uses self-attention for enhancing past estimates on the basis of the current estimate through backward flow of useful information shared across multiple audio segments. 

See Sec.~\ref{sec:supp_noise}-\ref{sec:supp_num_sources} for more analysis showing our model's separation superiority with 1) microphone noise, 2) varying inter-source distances, and 3) $k=3$
sources.

\paragraph{Qualitative Results.}
\begin{figure}[t]
\centering 
\includegraphics[width=0.75\linewidth]{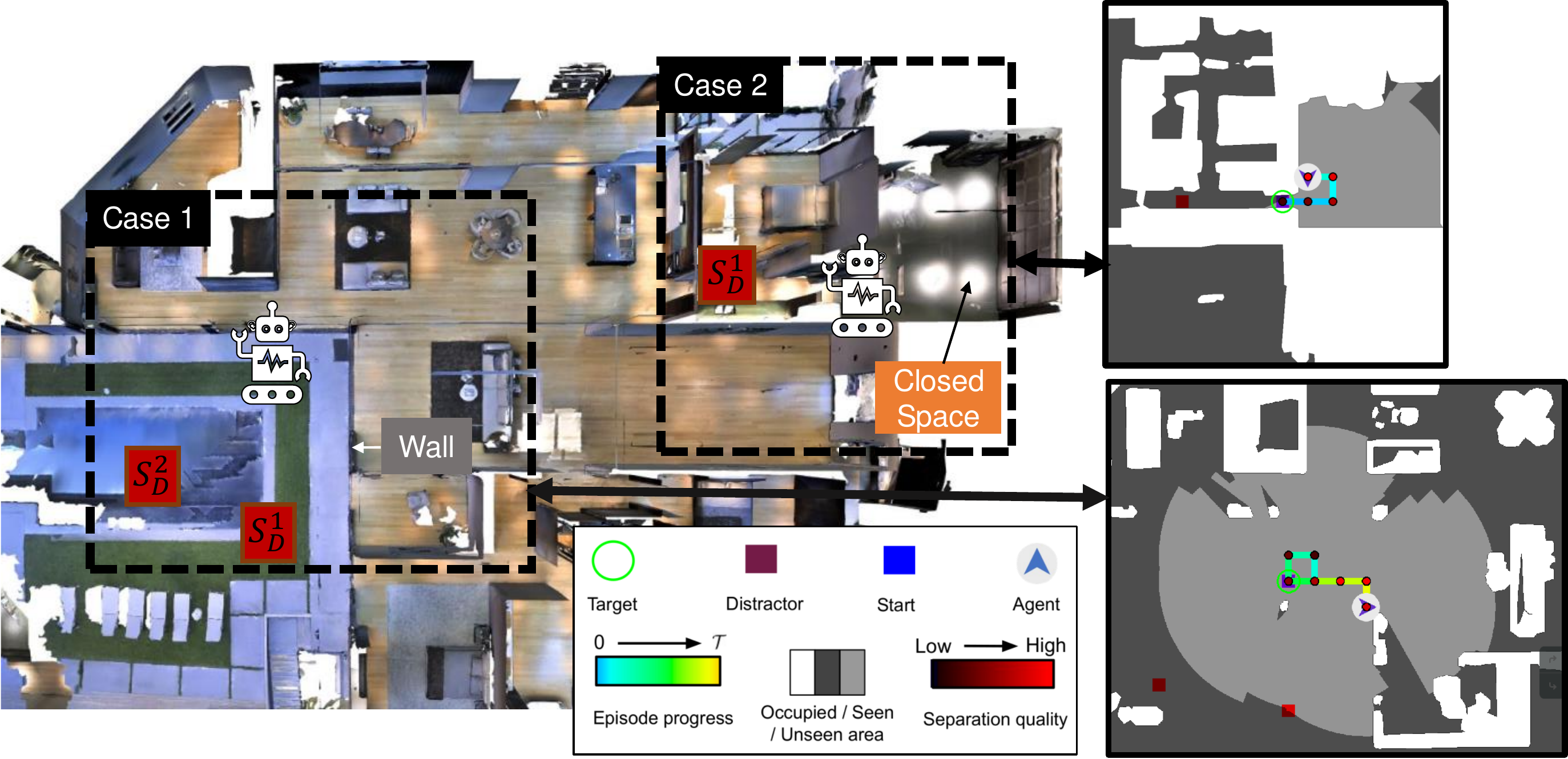} 
\caption{Sample trajectories of our model. Our model not only uses audio-visual cues to find locations in the 3D scene that help 
dampen the distractors, but also utilizes its transformer memory to provide separation robustness and stability, while non-myopically looking for sweet spots with 
the potential to improve overall separation in the future.}
\label{fig:qual}
\end{figure}

In Fig.~\ref{fig:qual}, we show two successful dynamic separation episodes of our method. In case 1, the agent starts outdoors and its motion 
choices are limited by the lack of potential acoustic barriers between the two distractors and the agent. Hence, it ventures away from the target in order to
go indoors and make use of a wall to place itself in the acoustic shadow of the distractors, while still being able to sample the target audio. Such behavior is possible with our transformer memory, which maintains reasonable separation quality while the agent crosses regions of low separation quality looking for places that could improve future separation. In case 2, the agent makes short movements to find a closed space in its vicinity that not only blocks the distractor, but also reinforces its hearing of the target through acoustic reflections. 
See Sec.~\ref{sec:supp_failure} and~\ref{sec:supp_sepHeatmap} for a discussion on our method's failure cases and our model's separation quality as a function of agent pose, respectively.

%% file: sections/conclusion.tex
\section{Conclusion}\label{sec:conclusion}
We introduced the task of active audio-visual separation of dynamic sources, where an agent must see and hear to move in an environment to continuously isolate the sounds of the time-varying source of interest. Our proposed approach outperforms the previous state-of-the-art in active audio-visual separation, as well as strong baselines and a popular exploration policy from prior work. In future work, we aim to extend our model to tackle  sim2real transfer and moving sources, e.g., with audio source motion anticipation.

\noindent 
\small{\textbf{Acknowledgements:} Thank you to Ziad Al-Halah for very valuable discussions. Thanks to Tushar Nagarajan, Kumar Ashutosh, and David Harwarth for feedback on paper drafts. UT Austin 
is supported in part by DARPA L2M, NSF CCRI, and the IFML NSF AI Institute. K.G. is paid as a research scientist by Meta.}

%% file: sections/supp.tex
\section{Supplementary Material}
In this supplementary material we provide additional details about:
\begin{itemize}
    \item Video (with audio) for qualitative depiction of our task and qualitative evaluation of our agent's performance (Sec.~\ref{sec:supp_video}).
    \item Failure cases of our method (Sec.~\ref{sec:supp_failure}), as referenced in Sec.~\ref{subsec:model_analysis} in the main paper.
    \item Separation quality of our method as a function of agent location in the 3D environment (Sec.~\ref{sec:supp_sepHeatmap}), as referenced in Sec.~\ref{subsec:model_analysis} in the main paper.
    \item Experiment to analyze the effect of our multi-step prediction parameter $E$ (
    Sec.~\ref{approach:audio_net} in main) on the dynamic separation quality (Sec.~\ref{sec:supp_E_effect}), as mentioned in 
    Sec.~\ref{sec:experiments} of the main paper.
    \item Experiment to gauge robustness of our method to audio noise (Sec.~\ref{sec:supp_noise}), as referenced in 
    Sec.~\ref{subsec:model_analysis} of the main paper.
    \item Experiment to show the dependence of dynamic separation quality on the number of audio sources (Sec.~\ref{sec:supp_num_sources}), as noted in 
    Sec.~\ref{subsec:model_analysis} of the main paper. 
    \item Experiment to show the dependence of dynamic separation quality on the minimum inter-source distance (Sec.~\ref{sec:supp_intersource_dist}), as referenced in Sec.~\ref{subsec:model_analysis} of the main paper. 
    \item Audio data details (Sec.~\ref{sec:supp_audio_data}), as mentioned in 
    Sec.~\ref{sec:experiments} of the main paper.
    \item Baseline details for reproducibility (Sec.~\ref{sec:supp_baselines}), as noted in Sec.~\ref{sec:experiments} of the main paper.
    \item Model architectures details for our method and baselines (Sec.~\ref{sec:supp_arch}), as noted in Sec.~\ref{sec:experiments} of the main paper.
    \item Training hyperparameters (Sec.~\ref{sec:supp_train_hyperparams}), as referenced in Sec.~\ref{sec:experiments} of the main paper.
    \item Definition of metrics used for measuring separation quality (Sec.~\ref{sec:supp_sep_metrics}), as mentioned in Sec.~\ref{sec:experiments} of the main paper.
\end{itemize}

\subsection{Qualitative Video}\label{sec:supp_video}
The supplementary video, available at \url{https://vision.cs.utexas.edu/projects/active-av-dynamic-separation/}, demonstrates the SoundSpaces~\cite{chen2020soundspaces} audio simulation platform that we use for our experiments, 
provides a qualitative depiction of our task, Active Audio-Visual Separation of Dynamic Sound Sources, and also illustrates the technical contributions of our approach over Move2Hear~\cite{majumder2021move2hear}, a state-of-the-art method for active audio-visual separation of static sounds. Moreover, we qualitatively compare our method with the best-performing heuristical baseline, namely Proximity Prior (Sec.~\ref{sec:experiments} in main), and Move2Hear~\cite{majumder2021move2hear} (Sec.~\ref{sec:experiments} in main), as well as qualitatively analyze failure cases. Please use headphones to hear the spatial audio correctly. 

\subsection{Failure Cases.}\label{sec:supp_failure}
Our agent fails to separate well when it moves too far away from the target in search of separation-friendly spots and hence cannot track the target after a point, even with the help of its transformer memory. Other failure cases involve the agent being stranded among multiple close-by audio distractors, while having very limited scope of movement due to the surrounding 3D structure. For examples of failure episodes, watch our qualitative video (Sec.~\ref{sec:supp_video}).

\subsection{Separation Quality vs. Agent Location}\label{sec:supp_sepHeatmap}
\begin{figure}[t]
\centering
    \includegraphics[width=0.45\linewidth]{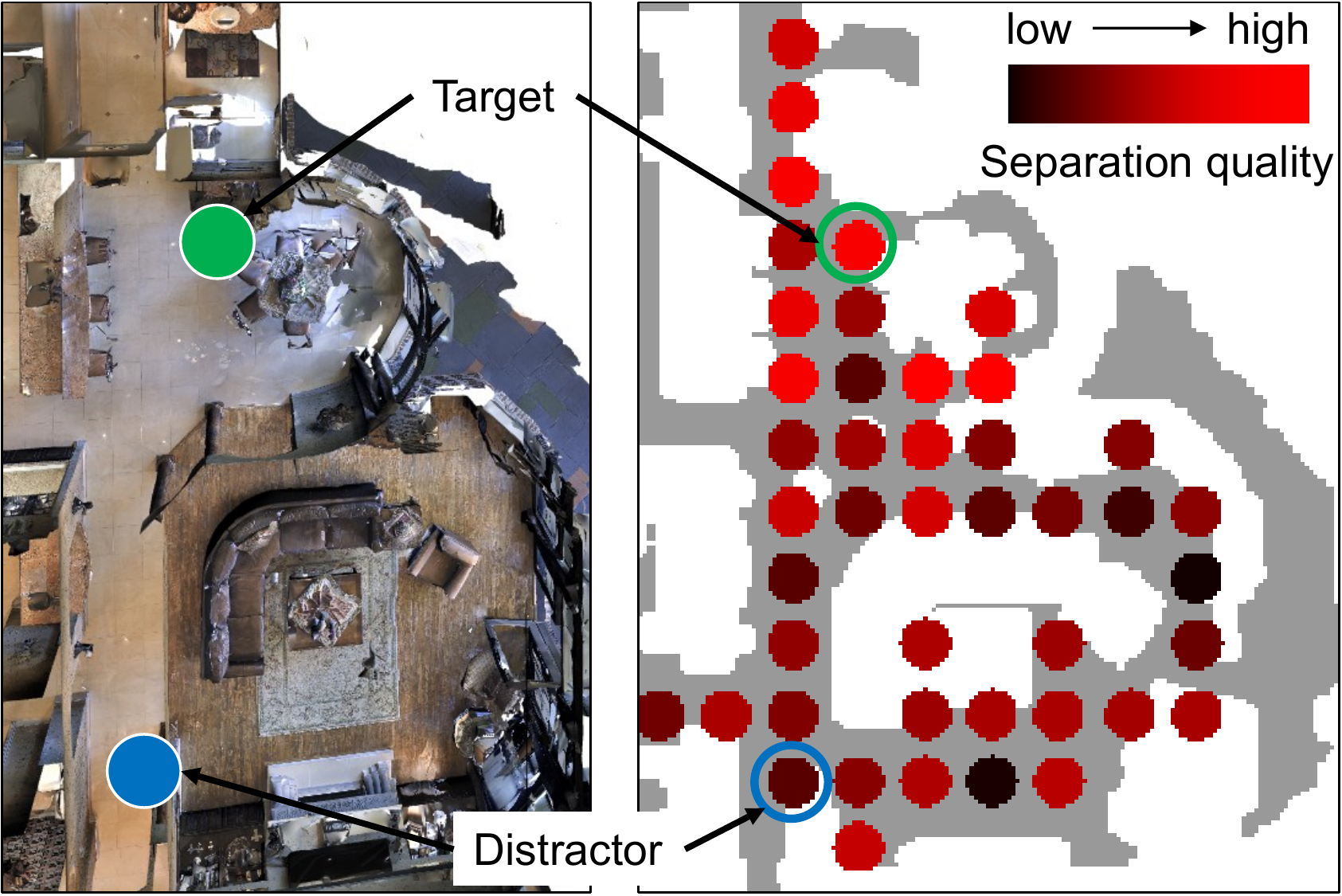}
\caption{Separation quality of our model when placed at different locations in a 3D scene for a given target source and distractor.
}
\label{fig:sep_heatmap}
\end{figure}

We show how our agent's separation quality evolves during its trajectory as a function of the observation poses it chooses, in the form of heatmaps in Fig. 4 in main and Supp video (4:57 - 5:19; 6:57 - 7:19). The fill color of a circle on the trajectory indicates its separation quality. Additionally, if we place our model at fixed locations and measure the separation quality (i.e., non active setting), we can produce a heatmap like Fig.~\ref{fig:sep_heatmap} where we can see how separation quality varies as a function of scene geometry.

\begin{figure}[t]
\centering 
\includegraphics[width=0.45\linewidth]{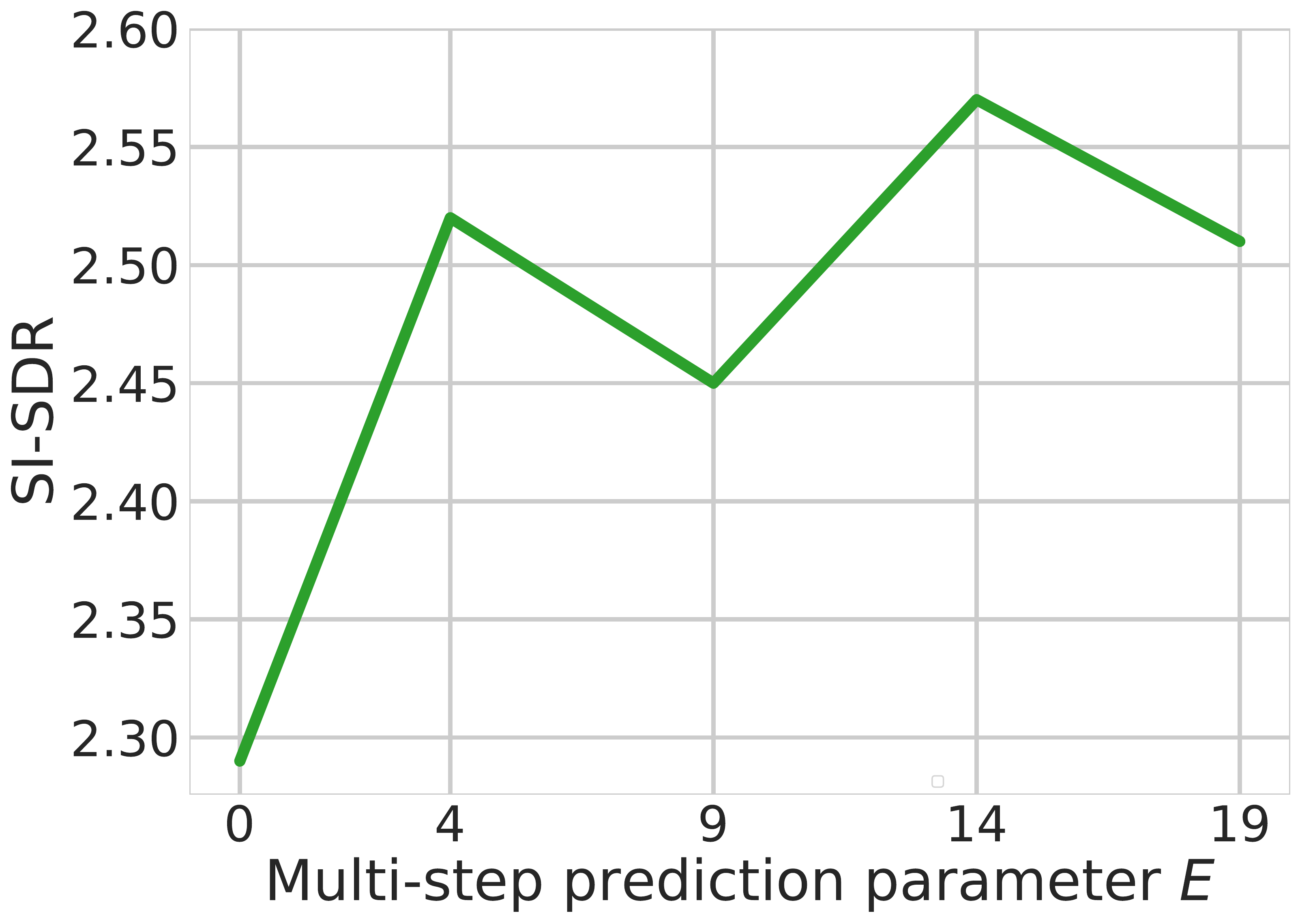}
\caption{Effect of our multi-step prediction parameter $E$ on dynamic separation quality.  Higher SI-SDR is better.}
\label{fig:e_val}
\end{figure}

\subsection{Multi-step prediction parameter $E$}\label{sec:supp_E_effect}
On setting our multi-step prediction parameter $E$ (Sec.~\ref{approach:audio_net} in main) to non-zero values other than 14 (Sec.~\ref{sec:experiments} in main), the overall separation quality (SI-SDR) on \emph{unheard} sounds drops up to 6.6\%. See Fig.~\ref{fig:e_val}. Lower $E$ values reduce our model's ability to improve past separations given the new observations. However, on increasing $E$ beyond a certain level, we see a degradation in separation quality as well. We expect that when the predictions are temporally distant, they are likely to be dissimilar in nature and their quality gets adversely affected due to the lack of useful cues in the current estimate for backward transfer.

\begin{figure}[t]
\centering 
\includegraphics[width=0.45\linewidth]{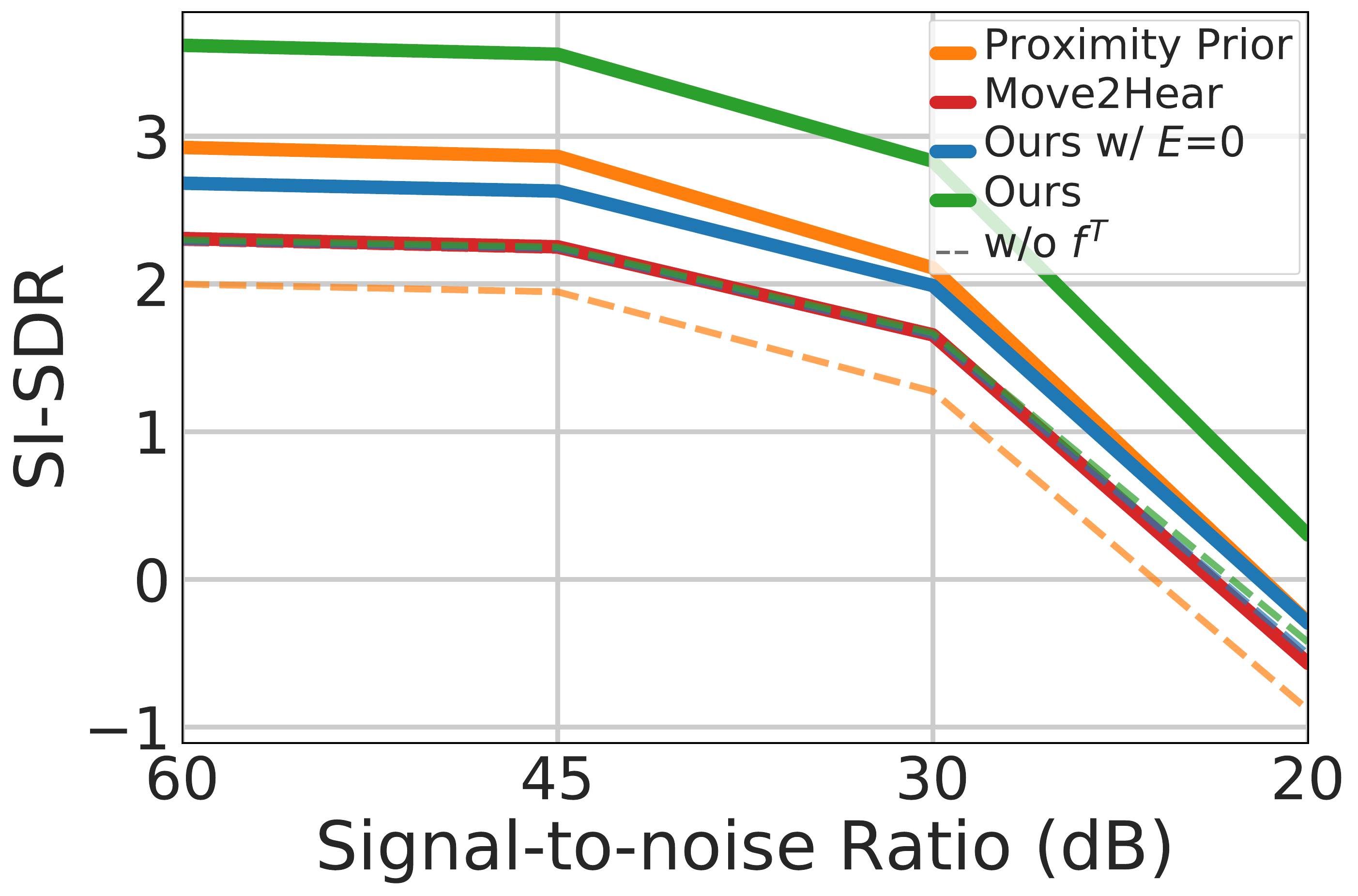}
\caption{Models' robustness to various noise levels in audio.  Higher SI-SDR is better.}
\label{fig:noise}
\end{figure}

\subsection{Audio Noise}\label{sec:supp_noise}
We test our method's dynamic separation performance in the presence of standard microphone noise~\cite{published_papers/29504406,takeda2017unsupervised} (Sec.~\ref{subsec:model_analysis} in main). See Fig.~\ref{fig:noise}. Our method robustly holds its superiority in separation over strong baseline models even at high noise levels (e.g., SNR of 20 dB, 30 dB, etc.~\cite{chen2020soundspaces}) 
See and hear our supplementary video (Sec.~\ref{sec:supp_video}) to get a better understanding of the distortion caused by the maximum noise that we evaluate, \ie SNR = 20 dB, where we play some noisy mixed binaural samples, as heard from the agent’s initial pose. Our method benefits from having the transformer memory $f^T$ (Sec.~\ref{approach:audio_net} in main), without which its separation performance drops sharply. Further, our multi-step prediction mechanism (Sec.~\ref{approach:audio_net} in main) for improving older separations in addition to separating the dynamic audio target for the current step provides a substantial boost in performance across all noise levels (compare the green and blue curves).

\subsection{Number of Audio Sources}\label{sec:supp_num_sources}

\begin{figure}[t]
\centering 
\includegraphics[width=0.45\linewidth]{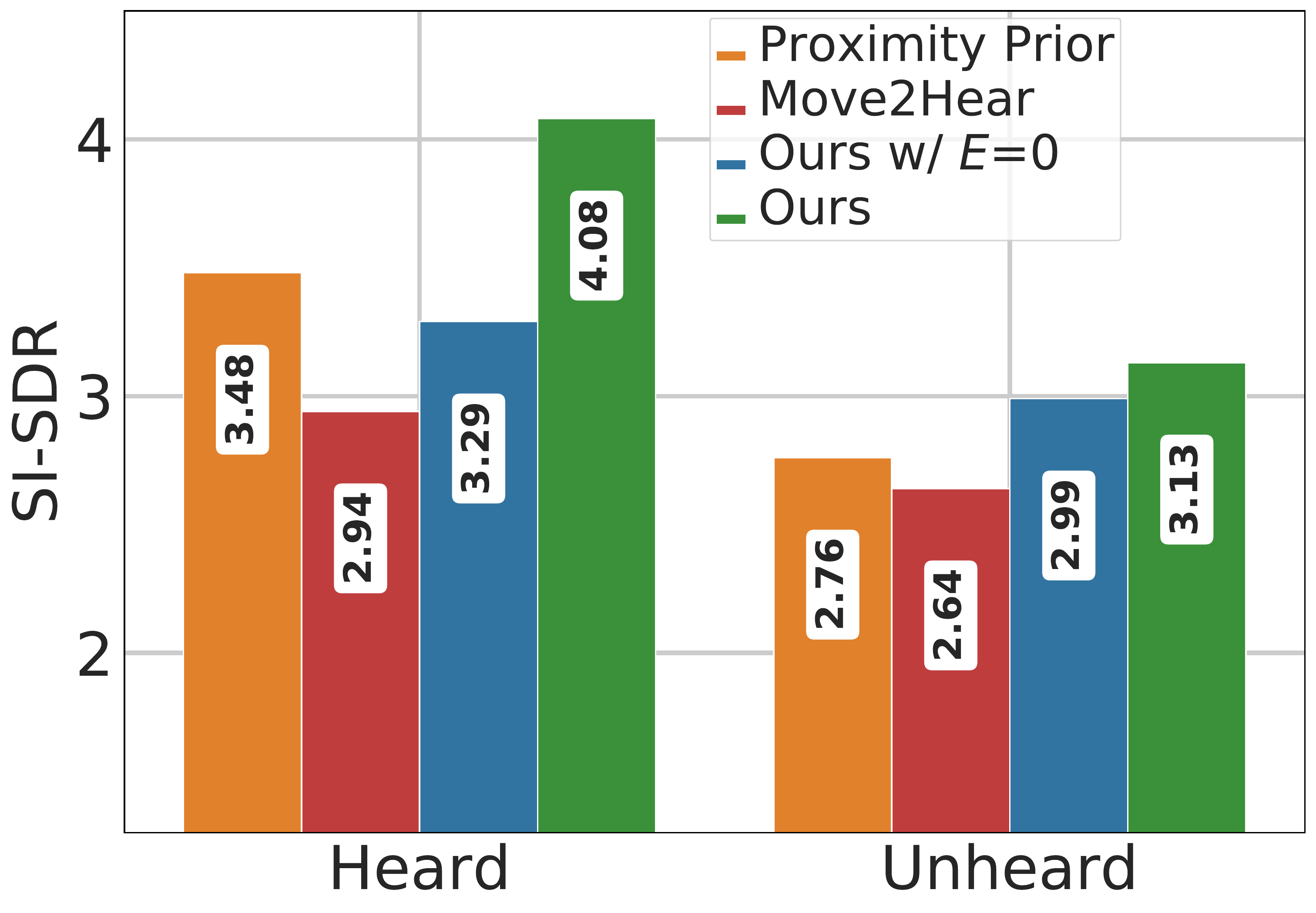}
\caption{Dynamic separation quality with 3 sources, \ie, 2 distractor sounds. Higher SI-SDR is better.}
\label{fig:3_src}
\end{figure}

We evaluate how increasing the number of dynamic distractor sounds in the environment affects separation performance (Sec.~\ref{subsec:model_analysis} in main). Fig.~\ref{fig:3_src} shows that even with $k=3$ sources (\ie, 1 target and 2 distractors) in every episode, our model leverages its smart motion policy and transformer memory to generalize better than the strongest of baselines. Further, multi-step predictions help tackle more distractors with both \emph{heard} and \emph{unheard} sounds (compare blue and green bars).

\subsection{Minimum Inter-Source Distance}\label{sec:supp_intersource_dist}

\begin{figure}[t]
\centering 
\includegraphics[width=0.45\linewidth]{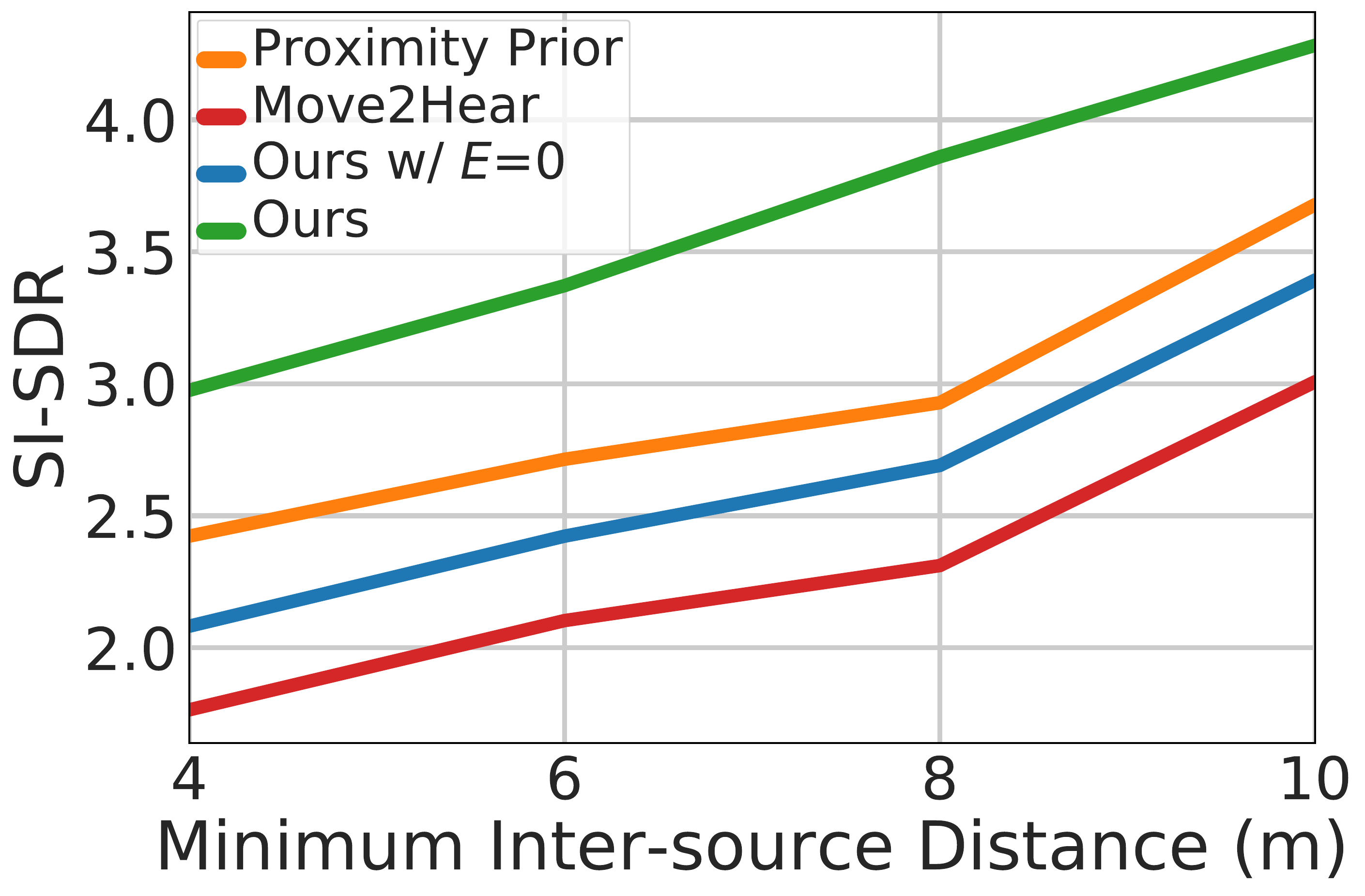}
\caption{Dynamic separation quality with minimum inter-source distance. Higher SI-SDR is better.}
\label{fig:s2s_hrd}
\end{figure}

We examine how changing the minimum inter-source distance for every episode in our dynamic audio setting affects the separation performance (Sec.~\ref{subsec:model_analysis} in main). See Fig.~\ref{fig:s2s_hrd}. Our method is able to maintain its performance gain over the most competitive baselines and also its own ablated version that makes single-step predictions. This is true even at very low values of inter-source distance, where our agent is severely cramped for room to move in search of separation-friendly spots, due to the close proximity to distractor sounds. This highlights the robustness to variations in the relative spatial arrangement of the target and distractor sources, which we can attribute to our method's joint learning of a dynamic separation policy and a transformer memory for making multi-step predictions for past and current targets.

\subsection{Audio Data}\label{sec:supp_audio_data}
Here, we elaborate on the details of the audio data that we use for our experiments (Sec.~\ref{sec:experiments} in main). 
\paragraph{Monaural Audio Dataset.}
Our monaural audio dataset contains 100 speaker classes from LibriSpeech~\cite{1164317}, 1 music class of different instruments from the MUSIC~\cite{zhao2018sound} dataset, and 1 class of assorted background sounds from ESC-50~\cite{piczak2015dataset} (Sec.~\ref{sec:experiments} in main). 

For LibriSpeech, each of the 100 speakers has at least 25 minutes of audio data in total. For all speech types and music, we join audio clips from the same type to form long audio clips, each of which is at least 20 s long, before splitting them into non-overlapping clips for train/val/test splits for unheard sounds (Sec.~\ref{sec:experiments} in main). For background sounds from our ESC-50 data, we replicate each 5 s clip four times and join them end to end to produce long clips of 20 s length. We choose replication for ESC-50 because joining distinct clips often doesn't lead to natural sounding audio due to high intra-class variance in the original dataset. 

We resample all clips at 16kHz and encode them using the standard 32-bit floating point format. Next, we compute the mean power across all clips and normalize each clip such that its power is equal to the pre-computed mean of 1.44. As a result, the mean ($\pm$std) audio amplitude in our dataset is 2.4 ($\pm$399.7)  for speech, -1.0 ($\pm$399.6) for music, and -0.7 ($\pm$399.6) for background sounds. The high std values for all sound categories indicate the high levels of dynamicity in all monaural clips, which not only make the dynamic separation task realistic but also challenging in nature.

Having preprocessed the long audio clips, we play a fresh audio segment at every dynamic source during an episode by sampling a random starting point within the long audio clips and shifting it forward by 1 s after every step. We loop over from the clip start if its end is reached during the course of an episode.

\paragraph{Spectrogram Computation.}
To compute spectrograms, we use the Short-Time Fourier Transform (STFT) with a Hann window of length 63.9ms, hop length of 32ms to promote significant temporal overlap of consecutive windows, and FFT size of 1023. This generates complex spectrograms of size $512 \times 32 \times C$, where $C$ is the number of channels in the source audio (1 for monaural and 2 for binaural). For all experiments, we use the magnitude of a complex spectrogram after reshaping it to $32 \times 32 \times 16C$, taking slices along the frequency dimension and concatenating them channel-wise to make model training computationally tractable. For all modules in our method and the baselines that take spectrograms as an input, except for the monaural audio encoder $\mathcal{F}^M$ (Sec.~\ref{approach:audio_net} in main) of the active motion policies, we compute the natural logarithm of the spectrograms  by adding 1 to all their elements for better contrast~\cite{gao2019visual-sound,gao2019co}, before feeding them to the respective modules.

Whenever the type label of the target audio source needs to be fed to a module along with a magnitude spectrogram, its value is looked up in a pre-computed dictionary with type names for keys and positive integers for label values, and concatenated with the input spectrogram after slicing.

\subsection{Baseline Details}\label{sec:supp_baselines}
We provide additional details about our baselines for reproducibility.
\begin{itemize}
    \item \textbf{DoA:} To face the direction of arrival (DoA) of the target audio, this agent first rotates clockwise from its starting pose until there is an adjacent node in front of it that's connected to the one it is currently at, then moves to the neighboring node along the connecting edge, and finally turns twice in the clockwise direction before holding its pose through the rest of episode for sampling direct sound from the target. 
    \item \textbf{Novelty~\cite{bellemare2016unifying}:} this agent is incentivized to maximize its coverage of novel states in its environment. In our setup, each node of the SoundSpace~\cite{chen2020soundspaces} grids is considered to be a unique state, which has an associated visitation count value that starts from 0 and is incremented every time the agent visits that state. At step $t$, the agent receives a reward:
    \begin{equation}
        r_t = \frac{1}{\sqrt{n_s}},
    \end{equation}
    where $n_s$ is the visitation count of its current state $s_t$.
    \item \textbf{Move2Hear~\cite{majumder2021move2hear}}: to account for dynamic audio, we modify the monaural audio encoder of its active audio-visual controller to only receive $\ddot{M}^G_t$ in place of the channel-wise concatenation of $\tilde{M}^G_t$ and $\ddot{M}^G_t$.
\end{itemize}

\subsection{Model Architectures}\label{sec:supp_arch}
\paragraph{Passive Audio Separator.}
The passive audio separator $f^P$ comprises a binaural extractor $f^B$ for extracting the target binaural given the mixed audio and a target type, and a monaural converter for predicting the target monaural from the extracted binaural (Sec.~\ref{approach:audio_net} in main).

$f^B$ and $f^M$ are U-Nets~\cite{RFB15a} (Sec.~\ref{approach:audio_net} in main). Their encoder is made of 5 convolution layers, each with a kernel size of 4, a stride of 2, a padding of 1 and a leaky ReLU~\cite{DBLP:conf/icml/NairH10,sun2015deeply} activation with a negative slope of 0.2. The number of convolution output channels are [64, 128, 256, 512, 512], respectively. Their decoder consists of 5 transpose convolution layers, each with a kernel size of 4, a stride of 2, a padding of 1 and a ReLU~\cite{DBLP:conf/icml/NairH10,sun2015deeply} activation. We append a convolution layer with a kernel size of 1 and stride of 1 to the U-Nets to produce the final spectrogram output of the networks.

\paragraph{Transformer Memory.}
Our transformer memory is a transformer encoder~\cite{vaswani2017attention} with 2 layers, 8 attention heads, a hidden size of 1024 and ReLU~\cite{DBLP:conf/icml/NairH10,sun2015deeply} activations. It has a pre-norm architecture that's been found to be well-suited for audio separation~\cite{subakan2021attention,zhang2021transmask}. Instead of using LayerNorm~\cite{ba2016layer} on the additive skip connection output, as proposed in the original transformer design~\cite{vaswani2017attention}, we use LayerNorm~\cite{ba2016layer} on the input side for both the multi-head attention and the feedforward blocks before the additive skip connection branches out. We use this block for every layer of the transformer encoder. 

We use a CNN for encoding the current and past monaural estimates $\tilde{M}^G$ from $f^P$ to 1024-dimensional features and add them with the corresponding sinusoidal positional encodings of the same dimensionality (Sec.~\ref{approach:audio_net} in main), before feeding them to the transformer encoder for self-attention. The encoder has 2 convolutions, each with a kernel size and a stride of 2, and 16 output channels. We use a ReLU activation~\cite{DBLP:conf/icml/NairH10,sun2015deeply} after the first convolution. For decoding the transformer encoder outputs to produce $\ddot{M}^G$s, we use another CNN with 2 transpose convolutions, each with a kernel size and a stride of 2, 16 output channels and a ReLU activation~\cite{DBLP:conf/icml/NairH10,sun2015deeply} (Sec.~\ref{approach:audio_net} in main). The decoder also receives the output of the first convolutional layer in the encoder as a skip connection, and adds it to the output from its own first transpose convolutional layer, before passing it to the second transpose convolution.

\paragraph{Observation Encoders.}
The visual encoder $\mathcal{F}^V$ (Sec.~\ref{approach:policy} in main) of our method and all baselines that uses one, namely Novelty~\cite{bellemare2016unifying} and Move2Hear~\cite{majumder2021move2hear}, is a CNN with 3 convolution layers with ReLU~\cite{DBLP:conf/icml/NairH10,sun2015deeply} activations, where the kernel sizes are [8, 4, 3], the strides are [4, 2, 1] and the number of output channels are [32, 64, 32], respectively. The convolution layers are followed by 1 fully connected layer with 512 output units. 

We use the same architecture as $\mathcal{F}^V$ for $\mathcal{F}^B$ and $\mathcal{F}^M$ (Sec.~\ref{approach:policy} in main), except that we use a kernel size of 2 instead of 3 for the last convolution. 

\paragraph{Policy Network.}
The policy network (Sec.~\ref{approach:policy} in main) for our method and the baselines with RL motion policies (\ie, Novelty~\cite{bellemare2016unifying} and Move2Hear~\cite{majumder2021move2hear}), comprises a one-layer bidirectional GRU~\cite{NIPS2015_b618c321} with 512 hidden units, and one fully-connected layer for its actor and critic networks.
\\\\We use He-normal~\cite{He_2015_ICCV} weight initialization for all network layers, except for the policy network GRUs, where we use semi-orthogonal weight initialization~\cite{saxe2013exact},  and the transformer encoder, for which we use the Xavier-uniform~\cite{pmlr-v9-glorot10a} initialization strategy.

\subsection{Training Hyperparameters}\label{sec:supp_train_hyperparams}
We pretrain $f^P$ by creating a static dataset of randomly sampled data points (Sec.~\ref{approach:training} in main), where each scene contributes a maximum of 30K data points, and using the Adam~\cite{kingma2014adam} optimizer with an initial learning rate of $5e^{-4}$ and a maximum gradient norm of $0.8$ until convergence.

We train the active motion policies of our method, Move2Hear~\cite{majumder2021move2hear} and Novelty~\cite{bellemare2016unifying} for 150 million steps with Decentralized Distributed PPO (DD-PPO)~\cite{wijmans2019dd}, where the weights on the value and entropy loss are 0.5 and 0.01, respectively, and the Adam~\cite{kingma2014adam} optimizer with an initial learning of $1e-4$ and a maximum gradient norm of $0.5$. We update the policy parameters after every 20 steps of agent's experience for 4 epochs. 

To jointly train $f^T$ or the acoustic memory refiner of Move2Hear~\cite{majumder2021move2hear} with the corresponding active motion policy, we use Adam~\cite{kingma2014adam} with an initial learning rate of $5e^{-3}$.

\subsection{Separation Quality Metrics}\label{sec:supp_sep_metrics}
Here, we provide additional details about our metrics for evaluating dynamic separation (Sec.~\ref{sec:experiments} in main).

\begin{enumerate}
    \item \textbf{STFT distance --} The Euclidean distance between the complex spectrograms for the monaural prediction and the ground truth,
    \begin{equation*}
        \mathcal{D}_{\{STFT\}} = ||\boldsymbol{\ddot{M}}^{G} - \boldsymbol{M}^{G}||_{2}.
    \end{equation*}
    \item \textbf{SI-SDR~\cite{8683855} --} We adopt an efficient nussl~\cite{nussl} implementation to compute the scale-invariant source-to-distortion ratio (SI-SDR) of a predicted monaural waveform in dB.
\end{enumerate}